%% file: main.tex
\documentclass[10pt, conference, compsocconf]{IEEEtran}
\usepackage[pdftex]{graphicx}
\usepackage[cmex10]{amsmath}
\usepackage{subcaption}     
\usepackage{array}
\usepackage{multirow}
\graphicspath{ {images/} }
\usepackage{tabularx}
\usepackage{makecell}
\usepackage{stfloats}
\usepackage[numbers]{natbib}
\usepackage{verbatim}
\usepackage{macros}
\usepackage{xcolor}
\usepackage[switch]{lineno}

\begin{document}


\title{A Cost Efficient Approach to Correct OCR Errors in Large Document Collections}


\author{
\IEEEauthorblockN{Deepayan Das, Jerin Philip, Minesh Mathew and C. V. Jawahar}
\IEEEauthorblockA{Center for Visual Information Technology, IIIT Hyderabad, India.}
\IEEEauthorblockA{ \{jerin.philip, deepayan.das, minesh.mathew\}@research.iiit.ac.in, jawahar@iiit.ac.in}}

\maketitle


\begin{abstract}
\input{sections/0_abstract.tex}

\end{abstract}

\begin {IEEEkeywords}
\textsc{ocr}, Batch Correction, Clustering, Post-Processing
\end{IEEEkeywords}

\section{Introduction}
\label{section:introduction}
\input{sections/1_introduction.tex}

\subsection{Related Work}
\label{section:related-work}
\input{sections/2_related_works.tex}

\section{Cost Effective Correction}
\label{section:formulation}
\input{sections/3-0_methods.tex}

\section{Grouping Error Words}
\label{section:clustering-methods}
\input{sections/3-1_grouping-error-words.tex}

\section{Dataset and Evaluation Protocols}
\label{section:dataset-evals}
\input{sections/4_evaluation.tex}

\section{Experiments, Results and Discussion}
In this section we describe the various components of our proposed batch correction model. We briefly discuss adapting our cost formulation to account for impurities when clusters are not homogeneous. Results for both batch correction schemes on annotated data, followed by error analysis of our clustering algorithms is then presented. Finally, we illustrate the performance of our model on the large scale, partially annotated dataset.
\label{section:experiments}
\input{sections/5_experiments.tex}

\section{Conclusion}
\label{section:conclusion}

\input{sections/6_conclusion.tex}

\bibliographystyle{unsrtnat}
{
\footnotesize
\bibliography{main.bib}
}

\end{document}

%% file: sections/0_abstract.tex
Word error rate of an {\sc ocr} is often higher than its character error rate. This is specially true when {\sc ocr}s are designed by recognizing characters. High word accuracies are critical to tasks like creation of content in digital libraries and text-to-speech applications.
In order to detect and correct the misrecognised words, it is common for an {\sc ocr} module to employ a post-processor to further improve the word accuracy. However, conventional approaches to post-processing like looking up a dictionary or using a statistical language model ({\sc slm}), 
are still limited. 
In many such scenarios, it is often required to remove the outstanding errors manually.

We observe that the traditional post processing schemes look at error words sequentially, since {\sc ocr}s process documents one at a time.
We propose a cost efficient model to address the error words in batches rather than correcting them individually. We exploit the fact that a collection of documents, unlike a single document, has a structure leading to repetition of words. Such words, if efficiently grouped together and corrected as a whole can lead to significant reduction in the cost. Correction can be fully automatic or with a human in the loop. Towards this we employ a novel clustering scheme to obtain fairly homogeneous clusters.
We compare the performance of our model with various baseline approaches including the case where all the errors are removed by a human. We demonstrate the efficacy of our solution empirically by reporting more than $70\%$ reduction in the human effort with near perfect error correction. We validate our method on Books from multiple languages.

%% file: sections/1_introduction.tex
The past decade witnessed a growing interest towards the creation of huge digital libraries
by digitizing books~\cite{gutenberg, google}.
One of the crucial steps towards digitization involves the recognition and reconstruction of document image collection(s) using an  {\sc ocr}. The  recognition module in the context of digitizing collections of books could be considerably different from that of recognizing a single document  image~\cite{xiu2008whole}. 
In this work, we extend this
idea to error correction in document image collections.


\begin{figure}[t] 
\centering 
    \includegraphics[width=\columnwidth]{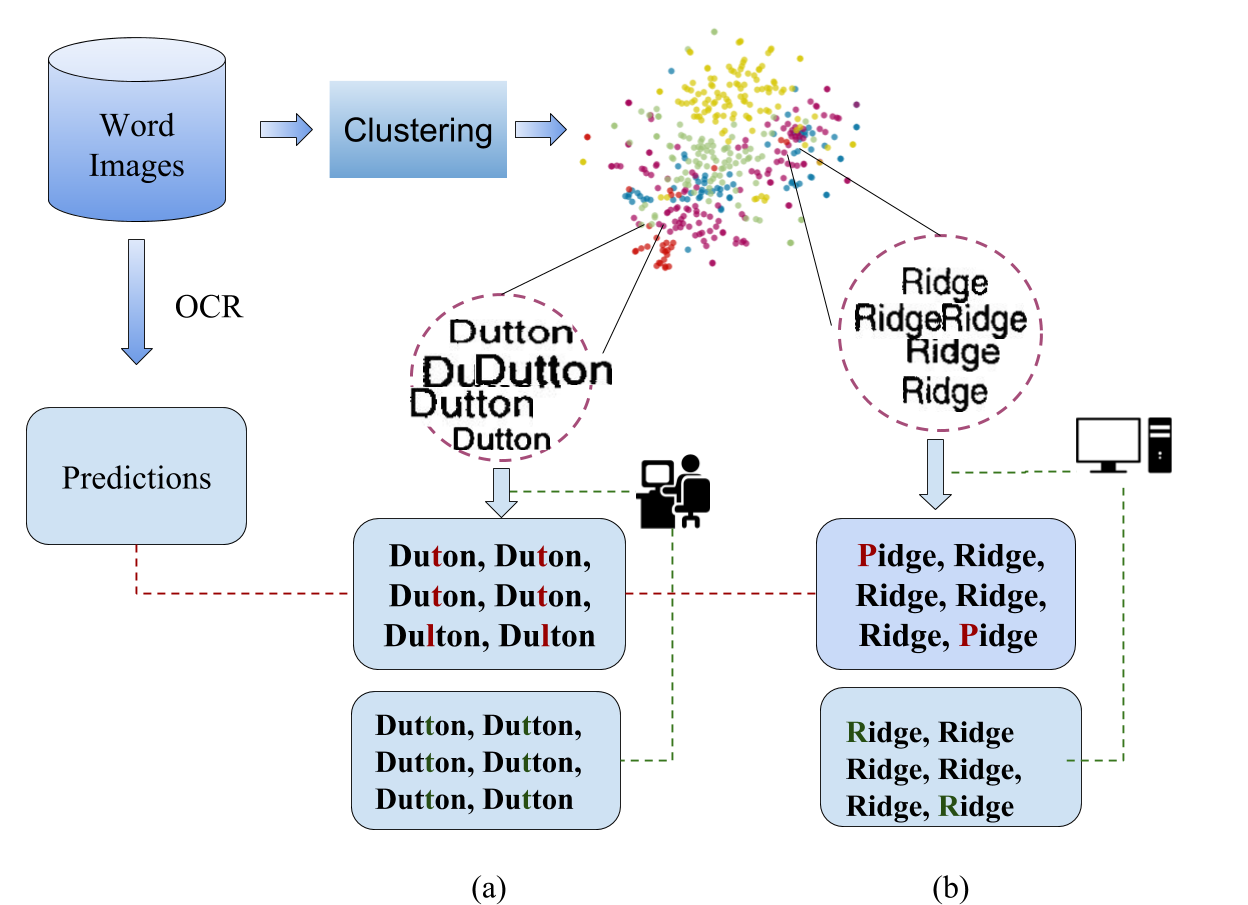}
    \caption{{\small The proposed pipeline for batch correction process where the error instances are clustered and corrected in one go. For a group of error instances, the correct label is chosen and applied. The correct label can be either chosen by a human annotator (a) or generated automatically (b).}}
    \label{fig:figure1} 
    \vspace{-0.7cm}
\end{figure}
 Often the recognition module of the {\sc ocr}s have an automatic error correction module embedded. This may be using a dictionary or a statistical language model ({\sc slm}). However, many applications need further improvement in accuracy.
 This demands a human intervention for removing these errors.
In this paper, we propose enhancements to the naive human correction approach which reduces the cost for human expert review by more than 70\%. Our work is guided by the following two insights. First - the {\sc ocr} module makes errors consistently. For two word images drawn from the same type of document, similar noise leads to the same kind of errors. We demonstrate this in Figure \ref{fig:consistent_errors} where instances of same word images drawn from a document collection are misclassified consistently by the {\sc ocr}. The second, there can only be a finite vocabulary for a book and majority of words unknown to the error detection system which may include named entities and domain specific terms repeat themselves throughout the collection. This is further validated in Figure \ref{fig:zipflaw} where we show that a subset of words in collection occur very frequently and constitutes almost $50$\% of the total words present. Under this setting, grouping based on image features or similarity in the predictions of the {\sc ocr} can provide cues for automatic correction or aide a human editor. We model the problem of error correction as batch correction where the human reviewer reviews and corrects errors in batches. Figure \ref{fig:figure1} presents an overview of our proposed batch correction scheme. Word image-prediction pairs extracted from a collection of documents form groups based on their image and text similarity. In case such a group is recognized incorrectly by the {\sc ocr}, only one instance from the group needs to be corrected which is then propagated to the rest of the group elements. Thus, correction needs to be made only once which reduces the cost of correction drastically. The correction can either be made with the help of a human editor or else the correction process can be automated.  We discuss both kinds of batch correction processes in detail later in this paper. The major contributions of this work are: 
\begin{itemize}
    \item We demonstrate how clustering can induce an automatic correction and reduce the manual effort in correction significantly.
    \item We successfully demonstrate ability to scale the clustering scheme to large collection of 100 books.
\end{itemize}

%% file: sections/2_related_works.tex

Conventional approaches to error detection and correction reduces to finding the closest match for an invalid word in a known vocabulary \cite{kukich1992techniques, postprocbassil2012}. 
 \citet{postprocbassil2012} put forth one of the first works which explored in detail {\sc ocr} post-processing methods, in which they consider three modes of correction. In the simplest of approaches, corrections could be performed manually by a human proofreader. Next, a dictionary-based method similar to what modern day word processors are equipped with was proposed. A possible correction is suggested once an error word was detected. This is accomplished by finding a word in the dictionary with minimum edit distance to the error word which becomes the correction proposal. Dictionary-based approaches could not capture errors in the grammar where words were correct according to the dictionary, but not in the surrounding context. Ability to correct such mismatches was brought about by grammar-aware models like Statistical Language Models using larger language context \cite{bassil2012ocr,saluja2017error}.
{\sc slm}s don't work well for many languages which lack corpus to train on. Also, they run into issues when newer out-of training domain words come in books. Further, \citet{smith2011limits} indicates unless carefully applied, a language model can do more harm than good. Hence it becomes necessary to review the results of a conventional {\sc ocr} system bringing a human in the loop for the perfect digital reproduction of a book.
To involve human in the loop, projects as early as Project Gutenberg \cite{gutenberg} introduced Distributed Proofreading \cite{newby2003distributed} approaches. Two proofreaders, having access to a book's page images, refine its {\sc ocr} outputs in turns. A demerit, in this case, is that the entire book has to be visited for proofreading. \citet{von2008recaptcha} using \textit{ReCaptcha} reports use of crowd-sourcing to transcribe word images where the {\sc ocr} outputs are detected to be erroneous. While the corrections are made only in the case of suspected errors, the efforts ignore the possibility of grouping similar misrecognized images and propagating the correct label to each instance in one go.
Observing {\sc ocr} errors to be highly correlated, \citet{abdulkader2009low} proposes a low-cost method to improve the required human-hours needed for correction using clustering. They group by {\sc ocr} outputs first, followed by finding subgroups using the word-image similarities. The above approach however, assumes the clusters to be completely homogeneous and thus fail to address cases where the clusters might contain more than one label.

\begin{figure}[h!] 
\centering 
 \includegraphics[width=\columnwidth]{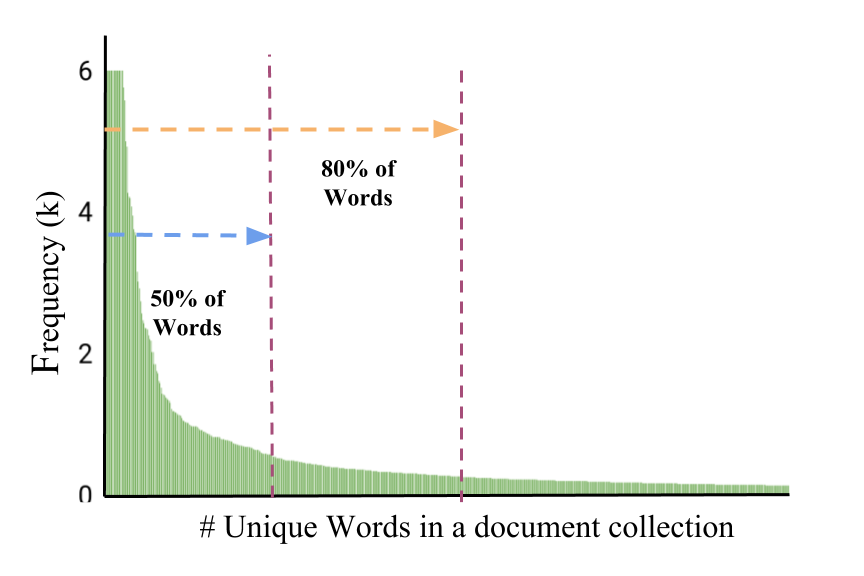}
    \caption{{\small The frequency of unique words in a collection of documents. A subset of words in the collection vocabulary have a very high frequency and accounting for 50\% of the words present in the collection. Thus it is safe to assume that if errors occurring in this subset are grouped and corrected in a batch, it can lead to a significant reduction in correction cost.}}
    \label{fig:zipflaw}
    \vspace{-0.5cm}
\end{figure}

In our next step, we review massive-digitization efforts in the past. Initiatives for a digital library for books through large scale digitization in the past include Project Gutenburg \cite{gutenberg}, Google Books \cite{google}. One of the main objectives of such projects is to provide content level access (enable search and retrieval) over the entire digitized collection.

\citet{baird2004document, taghva1996evaluation} note that enabling information retrieval in such databases is hampered by errors in {\sc ocr} outputs. Past works  turn to humans for correcting the last array of errors left in the pipeline post recognition \cite{von2008recaptcha,abdulkader2009low}. All these leave scope for improvement in the space of error correction, especially addressing challenges while scaling up the number of books. Our work is also motivated by the works of \citet{abdulkader2009low}. We group the errored predictions based on their image and text similarity and present them to a human editor. The human editor then decides the label for the cluster and also the components (elements present in a cluster) to which the label shall be assigned to. The instances where the cluster label do not match the content of the word image are addressed separately by the editor. This mitigates the propagation of errors for clusters that are not homogeneous.

\begin{figure*}[t] 
\centering 
 \includegraphics[width=\textwidth, height=180pt]{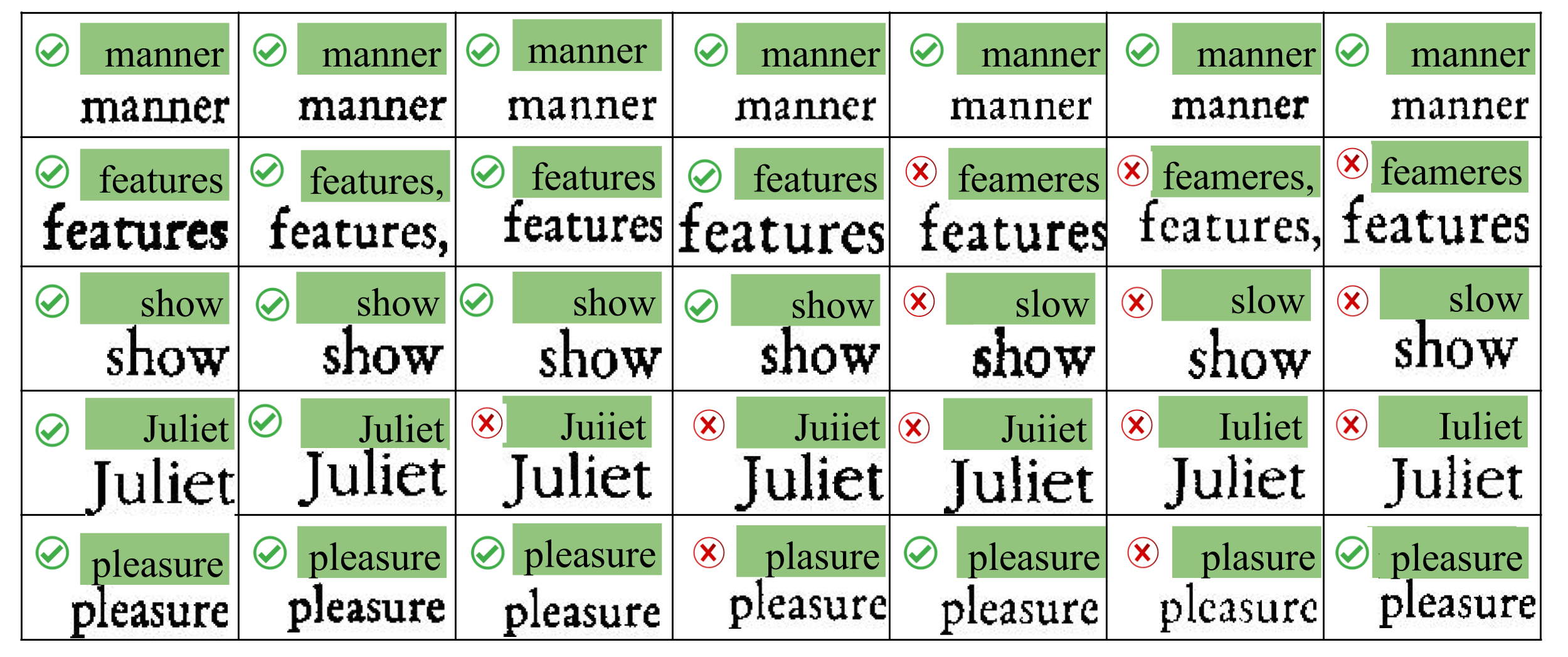}
    \caption{{\small Consistent errors generated by \textsc{ocr} for a given document collection. Each row represents different images for the same word and it's corresponding {\sc ocr} prediction in the green text box. We can observe that for similar degradations, the {\sc ocr} outputs similar error patterns.}}
    \vspace{-0.65cm}
    \label{fig:consistent_errors} 
\end{figure*}

%% file: sections/3-0_methods.tex

\label{subsection:cost}
In this section we formulate the problem of error correction and propose two strategies for using our batch correction method to address this issue.
\subsection{Problem Formulation}
Recognition modules of {\sc ocr} systems operate at a character or word level resulting in transcribing word-images into a textual string. Errors in such a setup are inevitable and the cost of manual correction is significantly high.
Since it is practically impossible to verify each word manually, we propose  to have an independent error detection mechanism operating on the {\sc ocr} predictions. Assuming that such a system has a low False Negative Rate, only instances where the {\sc ocr} prediction is not agreed upon by the error detection pipeline which we denote hereafter as \textit{error instances}, would then need to be corrected. We assume that the errors are detected with a dictionary or an appropriate error detection module. 
Our contribution is to make  further improvements to this setup by observing that an {\sc ocr} based system is prone to make systematic errors. Due to the nature of learning, multiple instances of the same word could be misclassified to the same wrong label. We propose a grouping of such misclassifications in a collection of documents which enable correcting these multiple errors in one go. In this work, we use a word-level {\sc ocr} and a dictionary for the error detection module.


One can categorize the agreement between the recognition module and the error detector into four:  

\begin{enumerate}
    \item Error False Positives ({\sc efp}): Words that are falsely flagged as error by the detection module since they do not exist in the dictionary {\sc oov}. 
    \item Error True Positives ({\sc etp}): Errors of the {\sc ocr} which are correctly detected by the error detection module.
    \item Recognizer False Negatives ({\sc rfn}): Words exist in the dictionary but are not the correct transcriptions of the word image.
    \item True Negatives ({\sc tn}) of the error detection module: Recognizer correctly predicts word image, and the detection module is in agreement.
\end{enumerate}




As far as the error correction is concerned, we would like to take human help or automatically correct the words categorized as {\sc etp}. Note that the words in {\sc tn} after error detection are correct words and nothing needs to be done. The words in {\sc rfn} cannot be detected as an error in isolation. Their correction needs larger language context and is out of scope for this paper. 
\begin{figure*}[t] 
\centering 
 \includegraphics[height=200pt]{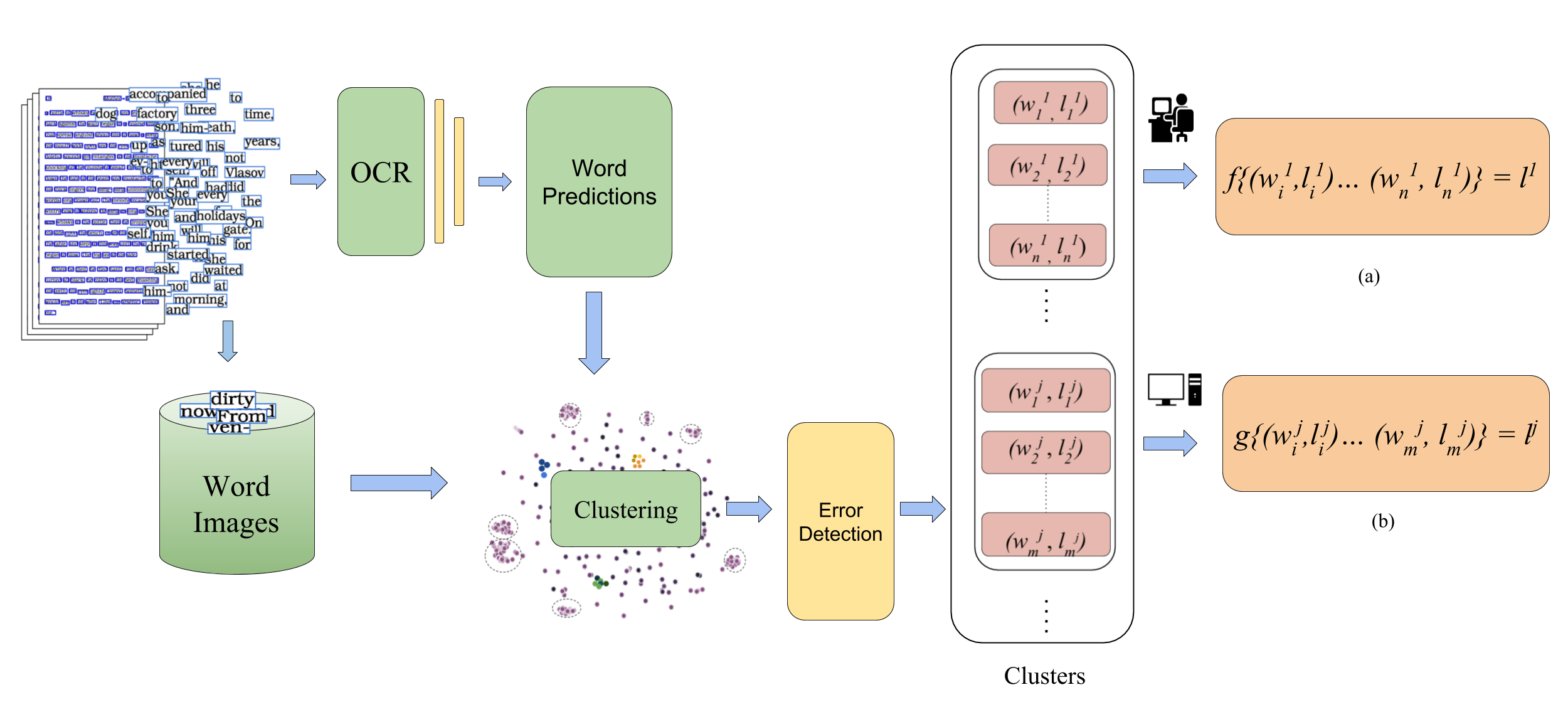}
    \caption{{\small Pipeline of proposed batch correction approach. Given word images ($w_{i}$\dots$w_{n}$) and its corresponding {\sc ocr} predictions ($l_{i}$\dots$l_{n}$) we form clusters. Next, the clusters containing error instances are sent for correction. We employ two forms of correction approaches which are shown in (a) when the human editor decides the label for a cluster and in (b) when the cluster label is generated automatically.}}
    \vspace{-0.65cm}
    \label{fig:pipeline} 
\end{figure*}

We propose a cost based evaluation to demonstrate the efficacy of our method. To this end, we first enumerate all possible edit actions a human in the loop has available and associate a cost with each action. We define a verification cost $C_v$ for the case where the reviewer just has to verify an already correct prediction to be a valid word. We define average word typing cost $C_t$ for cases where corrections have to be fully typed out. For cases where a dictionary provides correction proposals in a drop-down fashion, we define a cost $C_d$.

For a naive correction process (process where no batching is involved), the editor will have to type out corrections in {\sc etp} and verify a word wrongly classified in {\sc efp}. 
The total cost involved turns out to be $C_1 = (\lvert${\sc etp}$\rvert)C_{t} + (|${\sc efp}$|)C_{v}$. We denote this method hereafter as \textit{Typing}. If for some error instances, the editor has an additional option to select from a set of correction proposals, the cost reduces to $C_2 = |${\sc etp}$_t| C_t + |${\sc etp}$_d|C_d + |${\sc efp}$|C_v$ such that {\sc etp}$_t$, {\sc etp}$_d$ forms a partition of {\sc etp}. Here {\sc etp}$_t$ refers to the error true positives which can only be corrected via {\it Typing} whereas {\sc etp}$_d$ refers to the error true positives that can be corrected by choosing the correct suggestion from the set of correction proposals. (Here, $|X|$ denotes the cardinality of set $X$.) This method is denoted as \textit{Typing+Selection} hereafter.



We hypothesize correcting similar instances in {\sc efp} and {\sc etp} together can make digitization efforts more efficient. As mentioned above, we propose an approach that groups together error instances based on some similarity metric and propagate the correction of one of these to the rest of the group. We emphasize selection of the correction candidate for a group of error words can be either fully automated or done with human aid. We discuss both the propositions in detail later in this section. In the ideal case, word images with same ground truth will be grouped together, and the ability to correct them in one go would provide an efficient way for humans in the loop to correct large document collections. If we could group the error instances based on their ground truths as $C_1, C_2 \ldots C_{|V|}$, each of these groups could be corrected in just one action from editor leading to a cost of $ V_tC_t + V_dC_d + V_vC_v $ such that $V_t + V_d + V_v = |V|$. Here $V_t$, $V_d$ and $V_v$ are numbers of clusters requiring typing, selection from dictionary and verification respectively.

\subsection{Correction Approach}
\label{exemplar}

Our proposed model for error correction is presented in Figure \ref{fig:pipeline}. The document images, segmented at word level go through the {\sc ocr} pipeline which assigns them labels. The word images and their corresponding predictions are subsequently sent through a clustering pipeline, which groups the word images based on their image and text similarity. We discuss the clustering pipeline along with the features on which the clustering is performed in Section \ref{section:clustering-methods}. Next we perform an error detection on the components of each cluster and identify those clusters in which error instances occur. Only those clusters which contain error instances are sent for either of the two correction techniques- automated or human aided which are discussed below. 

\paragraph{Automated approach}
For a given cluster containing word images and their corresponding {\sc ocr} predictions, the most frequent prediction is chosen to be the representative of the whole cluster and its label is propagated to the remaining cluster elements. Two scenarios arise out of such a setting. For a given cluster-
\begin{enumerate}
    \item The number of correct predictions is more than incorrect predictions.
    \item The number of incorrect predictions is more than the number of correct predictions.
\end{enumerate}

In the first case, words appearing in {\sc etp} get corrected automatically without any further manual corrective action other than verification. In the second case, words appearing in {\sc efp} (proper nouns, acronyms, technical keyword, etc.) get corrected without much cost, while for clusters containing {\sc etp}, even the correct predictions end up being assigned the wrong label. Thus a human editor is required to verify the assigned label with the actual word image for every erroneous prediction and make keyboard entries wherever necessary. This leads to an added correction cost. 

\paragraph{Human aided approach} We allow a human editor to pick the representative of the cluster. This reduces the cost by eliminating the chances of error propagation which arise when labels are generated automatically. However, this also mandates that a human editor be present throughout the correction process. In case of {\sc etp}, the editor can enter the correction once and the correction is propagated to all matching images. Our method here reduces the cognitive load for the human, thereby improving efficiency.

In the above two approaches we consider the clusters to be completely homogeneous. Clusters containing impurities and the relevant correction approach is discussed later in the paper. 

%

%% file: sections/3-1_grouping-error-words.tex
In this section, we provide the details of our approach for grouping error words together.
As discussed earlier, this significantly reduces human cost.
\vspace{-0.15cm}
\subsection{Features for Clustering}
For every error instance, we have two types of features for use in clustering: text predictions of {\sc ocr} and features from word-image.

\paragraph{Image Features} We use the pre-final layer representations from deep neural networks trained to classify word-images. Such representations capture the discriminatory information between different word-images and have demonstrated success in embedding similar images together \cite{krishnan2018hwnet}. The activation for an image can be considered as a compact representation in a continuous space. For clustering the above features, we employ the \kmeans{} \cite{duda1973pattern} algorithm.

\paragraph{Text Features} For text features we propose using the word predictions of the {\sc ocr}. A natural distance measure for such features is the edit distance, which has been found to be of significant help for error detection in past work.
However, approaches like \kmeans{} are ill-suited to the discrete nature of these features and our distance measure. Therefore, we propose using a Minimum Spanning Tree ({\sc mst}) based approach \cite{duda1973pattern} using pairwise edit-distance to cluster variants of text predictions. This could also group consistent errors which comprise of error instances where the (1) Prediction is right but error detection is in disagreement. (2) Where for the same kind of word-image {\sc ocr} consistently give the same erroneous prediction due to bias in training data.

\paragraph{Image Features and Text} Word images with high visual similarity but having different text content can be grouped into the same cluster since they might be close to each other in the image feature space. This leads to fragmentation or formation of impure cluster. Assuming one true label per cluster can induce an additional cost of correcting word instances whose ground truth is different from the assigned label. To address the intra-cluster variability we further partition each cluster into sub-clusters by leveraging the textual transcription of each word image such that words that lie within a predefined edit distance, can be grouped into the same sub-cluster.

\subsection{Clustering Algorithms}

In a simpler first approach over a fewer number of books, we use \kmeans{} and {\sc mst} based clustering algorithm to group error instances together. While the two algorithms work well for a fewer number of books, they are not well suited to scale to a larger setting. We address this by using a Locality Sensitive Hashing ({\sc lsh}) based nearest neighbour computation \cite{indyk1998approximate} in our clustering pipeline. We discuss in detail the algorithms and their suitability below.

We employ \kmeans{} on the image representations with number of clusters ($k$) set to number of unique words in a collection. The \kmeans{} algorithm has a time complexity of $O(n^{2})$ where $n$ is the number of error instances detected by our pipeline. 

We use {\sc mst} clustering on the text predictions to further partition the clusters. We consider the predictions as vertices of a weighted undirected graph, and the pairwise edit-distance between two vertices form the edge weights. Distances between vertices are scaled to $[0, 1]$. 
A {\sc mst} is constructed and edges with weights greater than a threshold 
are discarded, which results in a forest where each connected component forms a cluster. 

Degradations in print, paper or both over time are prevalent in older documents. Font styles and variations different from the {\sc ocr}'s training distribution used by a common publishing system across these books could be similar in the image space. Similar noise in the images like the cuts and merges lead to consistent errors in {\sc ocr}. This prior domain knowledge can be incorporated and taken advantage of while clustering. 
Under these circumstances, we find {\sc lsh} well suited for scaling up correction in our problem setting. {\sc lsh} tends to approximate the nearest neighbour search in a way such that items which are similar are hashed into the same `bucket'. Consistency in noise leads to similar hashes for features from images with similar content. Search space is now limited to the bucket of word-images for which hash matches the query image. This makes the process orders faster.

%% file: sections/4_evaluation.tex
 Our dataset comprises of books that have been fully digitized by our {\sc ocr} module. They are categorized into two types. The first is a smaller subset having books that have been verified by a human expert, while the second composes a larger subset containing unverified books.
 We denote the former as \textit{fully annotated} data and the latter as \textit{partially annotated} data. 
 We seek to evaluate both these datasets on two separate objectives.
 
 For the \textit{fully annotated} dataset, our objective is to find which among the proposed clustering approaches works best for document collection. For the \textit{partially annotated} dataset, we look to evaluate the scalability of the proposed clustering approaches on larger unverified data. Table \ref{tab:dataset-descr} gives details of our dataset, both \textit{annotated} and \textit{partially annotated} used in our experiments and the evaluation methods directed towards ascertaining what works for each objective.
\begin{table}[ht]
\centering

\begin{tabular}{|c | l c c c c|}
\hline
scale &   language   & \#books & \#pages & \#words & \# unique \\ \hline
\multirow{2}{*}{FA} & English & 15 & 2417 & 0.73M & 30K \\
                        & Hindi & 32 & 4287 & 1.20M & 63K \\ \hline
\multirow{2}{*}{PA} & Hindi & & & & \\
                       &- annotated & 50 & 200 & 30K & 6K \\
                       &- unannotated &100 & 25K& 5M\textsuperscript{*}  & 80K\textsuperscript{*} \\
                       \hline
\end{tabular}
\caption{Details of the books used in our work. Here FA refers to the fully annotated books whereas PA refers to the partially annotated books.}
\vspace{-0.65cm}
\label{tab:dataset-descr}
\end{table}
\subsection{Annotated Data}
The annotated dataset comprises of 19 books from English and 32 books from Hindi. Pages from the books are segmented at a word level and annotated by human experts. 5 books are set aside from each of the languages to train the  {\sc ocr} while rest of the books are used for testing and further batch correction experiments. 
\subsection{Partially Annotated Data}
 In order to demonstrate the scalability of our approach, we run our experiments on a larger collection containing 100 Hindi books.  Most of these books were printed decades ago, resulting in degradation in quality of pages.  The collection consists of almost 25,000 pages with more than 5 million words. A subset of 200 pages across 50 books are set aside as test set, for which we obtain bounding boxes and ground truths annotated by human experts. 
 
 \input{tables/auto-results.tex}

 
 

\subsection{Evaluation on Fully Annotated Dataset}
We want to find how many time units would be saved for a human editor using our pipeline compared to the case where each error instance have to be visited individually. 
The cost is measured in units of seconds  of human effort put into correction. The following values are used for computing the cost in simulations. For verification cost $C_v$, we supply a 1 second and for picking a choice from suggestions, we set cost $C_d$ to be $5$ seconds. The cost of typing, $C_t$, is set to $15$ seconds, for each word. 

The numbers are compared across all proposed clustering approaches. Having fully annotated ground-truth information gives complete cost required in this setting.


\subsection{Evaluation on Partially Annotated Dataset}

Here, we evaluate the performance of our approach on a large collection of 25,000 pages. We estimate the performance on this collection by
explicitly measuring the performance on the test subset of 200 pages. Only this test set is used to infer performance, even though we run clustering on a larger set varying the size ranging from 200 to 25,000 pages. Please note that the performance reported in this collection is only an approximation.

We hypothesize that the increase in word accuracy translates to a reduction in correction cost. Also, since the subset of pages used in evaluation belong to the same pool of books which the larger clustering algorithm is run on, it is reasonable to assume that decrease in cost during evaluation is indicative of a decrease in the larger set of pages.

%% file: tables/auto-results.tex
\begin{table*}[t]
    \centering
    \begin{tabularx}{\textwidth}{|l | X X X | X X X || X X X | X X X|}
    \hline 
    & \multicolumn{6}{c ||}{English} & \multicolumn{6}{c|}{Hindi} \\\cline{2-13}
  Method           
    
    & \multicolumn{3}{c |}{Automated} & \multicolumn{3}{c ||}{ Human}
    & \multicolumn{3}{c |}{Automated} & \multicolumn{3}{c |}{ Human}
    \\
    & Typing & \multicolumn{2}{c |}{(Typing + Selection)}
    & Typing & \multicolumn{2}{c ||}{(Typing + Selection)}
    & Typing & \multicolumn{2}{c |}{(Typing + Selection)}
    & Typing  & \multicolumn{2}{c |}{(Typing + Selection)} \\
    & - & Static & Growing 
    & - & Static & Growing 
    & - & Static & Growing 
    & - & Static & Growing \\\hline
\kmeans(I)        
    & 1.130 & 0.873 & 0.692 
    & 0.689 & 0.527 & 0.372 
    & 1.013 & 0.714 & 0.648 
    & 0.494 & 0.366 & 0.234\\ 
{\sc lsh}(I)           
    & 0.939 & 0.732 & 0.695 
    & 0.283 & 0.232 & 0.222 
    & 0.944 & 0.664 & 0.659 
    & 0.162 & 0.135 & 0.134\\ 
{\sc mst}(T) &
   1.000 & 0.740 & 0.695 & 
   0.199 & 0.187 & \textbf{0.187}  & 
   1.000 & 0.695 & 0.681 &
    0.142 & 0.133 & 0.132 \\
\kmeans(I) + {\sc mst}(T) & 
    1.000 & 0.853 & \textbf{0.653} & 
    0.607 & 0.459 & 0.327 & 
    0.960 & 0.681 & \textbf{0.634} & 
    0.281 & 0.217 & 0.191\\ 
{\sc lsh}(I) + {\sc mst}(T)   
    & 0.947 & 0.739 & 0.689 
    & 0.285 & 0.232 & 0.222 
    & 0.949 & 0.666 & 0.651 
    & 0.153 & 0.129 & \textbf{0.128}\\ 
    \hline
    \end{tabularx}
    \caption{Evaluation of costs of each approach proposed in this paper. The numbers reported are relative to Full Typing method. We observe a decrease in cost as we go left to right for each clustering approach for books of a given language. `I' stands for image features, and `T' stands for prediction text. }
    \vspace{-0.5cm}
    \label{tab:comparison-relative-costs}
\end{table*}

%% file: sections/5_experiments.tex
\subsection{Our Pipeline}

Our setup consists of the following components - an {\sc ocr} module for recognizing word images, a Convolutional Neural Network({\sc cnn}) for extracting representations from these word images and an error detection module for verifying the accuracy of these predictions.
Our {\sc ocr} implementation follows a hybrid architecture with convolutional and recurrent layers first proposed by \citet{shi2017end} in their work towards scene-text recognition. 


We trained two {\sc ocr}s -- one for each language. For training, we set aside 5 books each from English and Hindi book datasets respectively. The English language {\sc ocr} was trained on word images from nearly 600 pages ($\sim$160K words) while the Hindi language  {\sc ocr} was trained on word images from approximately 650 pages ($\sim$180K words).

The {\sc cnn} based feature extractor used in our experiments follows the architecture described in \citet{krishnan2018hwnet}. The network was initially trained on synthetic handwritten word images and later fine-tuned on a real-world corpus. Real data used in training this network is the same as 160K word images which were used for training our {\sc ocr}. The segmented word-images are fed to the network and the pre-final layer activations are used as features for clustering.

The error detection module is realised by a dictionary. An instance is determined to be erroneous if its prediction is not present in the dictionary. To suggest corrections for an error instance, the dictionary requires a reasonably good vocabulary. We generate a base dictionary by using Wikipedia dumps for the respective language. For each book while testing, we enrich the corresponding base dictionary's vocabulary further using ground truths of books used for training but not the ones we are testing. 
We use two variants of this dictionary - one \textit{Static} and the other \textit{Growing}. The \textit{Growing} allows for addition of new words to dictionary, like how modern word processors do. In our grouped correction scenario same words could be scattered across clusters and \textit{Growing} dictionary speeds up correction by not having to type again the words already corrected.

\begin{figure*}[t] 
    \includegraphics[width=\textwidth]{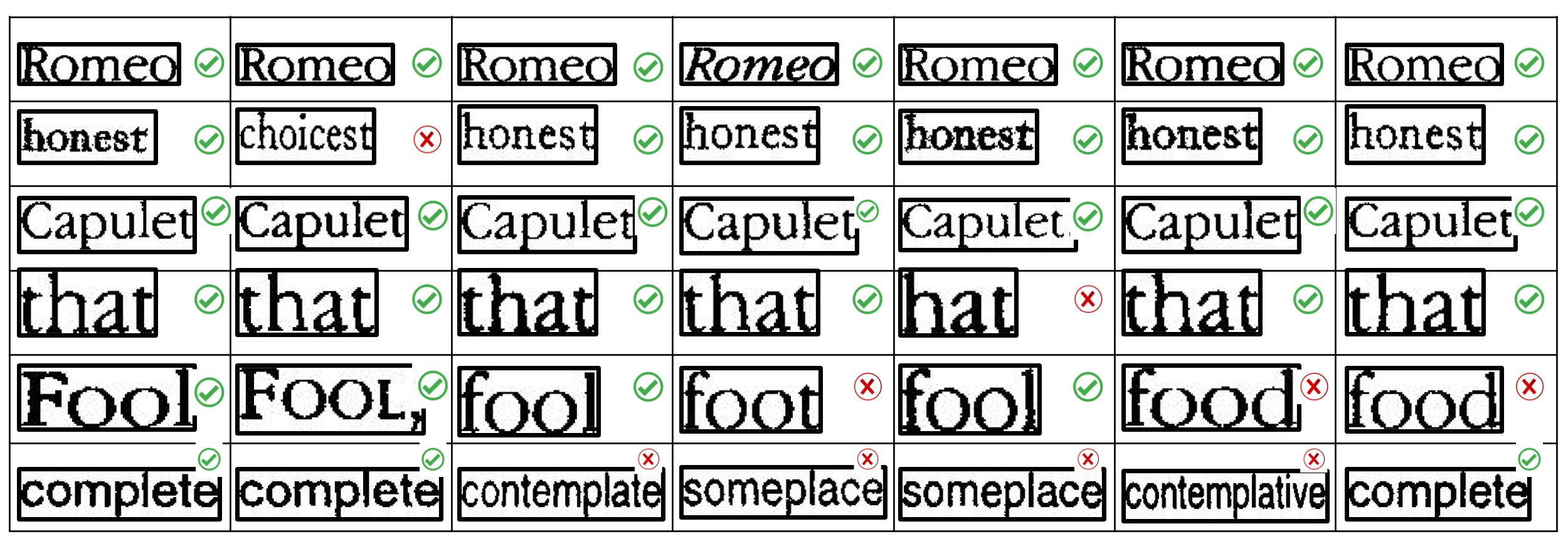}
    \caption{{\small Qualitative results of \kmeans{} + {\sc mst} clustering on English dataset. Images, relevant to the cluster are marked correct while the false positives are crossed out.}}
    \vspace{-0.65cm}
    \label{fig:QualRes} 
\end{figure*}

\subsection{Cluster Impurity}

One of the limitations of clustering algorithms like \kmeans{} or {\sc mst} is their inability to form completely homogeneous clusters. Despite our efforts in fusing image and text features together in order to minimize the impurities, still outliers manage to creep into the clusters. This can be verified in Figure \ref{fig:QualRes}. This poses a serious drawback in our error correction pipeline. Up until now we considered our clusters to be homogeneous and formulated our cost accordingly. However, in practice this can lead to wrong cost estimation. For automated approach, cluster impurity can lead to assignment of labels to instances which do not share the same ground truth. Thus an annotator needs to revisit each cluster and correct all unwarranted cluster assignments. 

For human in the loop, we let the human assign labels to the cluster components. A human can correct impure parts of the cluster by visual inspection through {\it Typing} or {\it Selection}. Consistent errors can be corrected for a group in this case, unlike in automated approach giving this method an advantage.

\subsection{Results and Discussions}
All costs in this work are computed relative to {\it Typing}. Table \ref{tab:baseline} delineates the cost for correction without grouping efforts. We experiment with setups involving no dictionary, static as well as growing dictionaries, restricting the edit actions available accordingly.  We find {\it Typing + Selection} outperforms {\it Typing} and {\it Growing} outperforms {\it Static}, as expected.

\input{tables/relative-cost-corrections.tex}

In Table \ref{tab:comparison-relative-costs}, we compare the cost of correction when we employ different clustering schemes. Here corrections are performed in batches. The rows correspond to clustering algorithms.
Our results are across the two correction approaches - the first which is automated and the second involving a human editor. 

The order among relative costs for edit actions and dictionary variants are consistent with the case without batch correction (Table \ref{tab:baseline}).  Further, we find sequential refinement of clusters using image features and then text-features perform best among different clustering schemas.  
For the automated approach, \kmeans{} on image features followed by {\sc mst} on text features achieve the lowest cost for both the languages. When involving human editor in the process, for Hindi, {\sc lsh} on image features and refining with {\sc mst} works best, while for English data {\sc mst} on text word predictions seems to achieve the lowest cost. 


Correction methods involving human editor consistently outperforms the automated correction approach, even with the former restricted in actions and in dictionary. This can be attributed to the failure of automatic approach in determining which is the correct prediction in a cluster which is largely impure. 



\subsection{Error Analysis}
\begin{figure}[h]
\centering
  \begin{subfigure}[b]{0.46\columnwidth}
    \includegraphics[width=\linewidth]{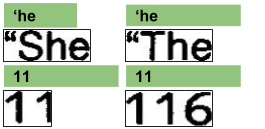}
    \caption{Text predictions}
    \label{fig:failure-cases-text}
  \end{subfigure}
  \begin{subfigure}[b]{0.46\columnwidth}
    \includegraphics[width=\linewidth]{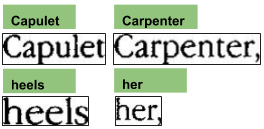}
    \caption{Image features}
    \label{fig:failure-cases-images}
  \end{subfigure}
  \caption{{\small Failure cases for clustering on text and image features respectively. Each row in the above figure represents one cluster. The text predictions are depicted in green text box.}}
  \label{fig:ErrorAnalysis}
  \vspace{-0.4cm}
\end{figure}

We discuss failure cases of our proposed correction process with a few qualitative examples. For text predictions clustered using {\sc mst} algorithm, a few error cases are illustrated in Figure \ref{fig:failure-cases-text}. The recognition module's high confusion in predicting numbers and punctuation extends to clustering using text predictions. But there exists strong cues here in the image feature space which can be used to group samples separately.

Figure \ref{fig:failure-cases-images} shows failures in clustering solely using image features. Instances containing `Carpenter' and `Capulet' are grouped into the same cluster although there is a significant difference between their text predictions. Image feature based clustering alone fails to obtain a pure cluster here, but text predictions' similarity can be used to make clusters more pure. We demonstrate such successful refinement in Figure \ref{fig:QualRes}. `Capulet' is one such correction proposal, but the entry corresponding to `Carpenter' is no longer associated. 

Failure cases of the combined clustering approach are indicated in Figure \ref{fig:QualRes}. Predictions `fool' and `food' are inherently different, but still managed to be clustered together. This is likely due to these being very near in image and text space.
\subsection{Results on Large Dataset}

\begin{figure}[h!] 
    \includegraphics[ height=150pt]{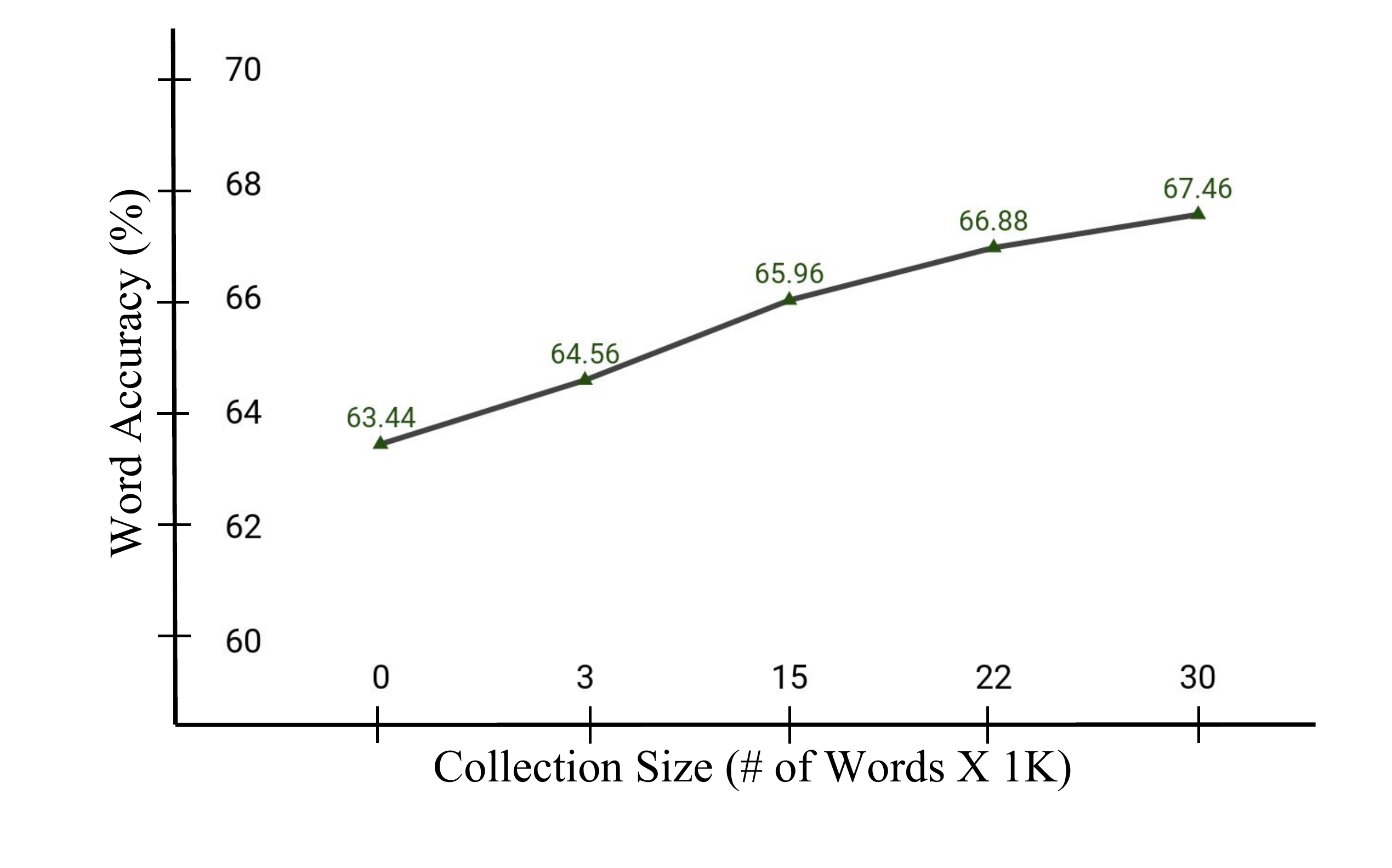}
    \caption{{\small Result on the unannotated data. We observe that as the number of words in the collection increases the automated batch correction method's ability to correct the errored predictions improve which is reflected by the increase in {\sc ocr} accuracy.}}
    \vspace{-0.3cm}
    \label{fig:LarRes}
\end{figure}

We vary the size of the collection from 200 to 25,000 pages and estimate the accuracy on the 200 fully annotated pages. 
Due to the nature of the word images, the word accuracy on the 200 annotated pages turns out to be quite low ($64\%$), which suggests that there is scope to improve the word accuracy using our batch correction techniques. Our main objective is to demonstrate that as we increase the collection size, our automated batch correction method becomes better at picking the right candidate. For this, we perform clustering on data of different sizes, where we keep on increasing the number of words for each subset. We observe from Figure~\ref{fig:LarRes} that the word accuracy for the dataset improves as the size of collection increases. This implies that for the larger unannotated data, the proposed batch correction method will lead to a better improvement in word accuracy and thus reduction in overall correction cost.

Performance of the traditional methods for error correction does not change with the size of the collection. Our method scales well to large collections and yields superior performance, making it an ideal candidate for large scale efforts like digital libraries.


%% file: tables/relative-cost-corrections.tex
\begin{table}[h!]
\centering
\begin{tabular}{|c|c|c|c|}
\hline
& Typing & \multicolumn{2}{c|}{Typing + Selection} \\\cline{2-4}
& - & Static & Growing \\\hline
English & 1.000  & 0.740 & \textbf{0.695} \\
Hindi   & 1.000  & 0.686 & \textbf{0.681} \\\hline
\end{tabular}
\caption{Relative cost of correction with respect to full typing when no batching is involved.}
\vspace{-0.55cm}
\label{tab:baseline}
\end{table}

%% file: sections/6_conclusion.tex
In this work we propose a cost efficient batch correction scheme involving a human editor. We also propose a novel clustering schema to improve the homogeneity of clusters which leads to significant reduction in correction cost. We compared our method with various baseline approaches.
We also demonstrate the scalability of our batch correction on a large digitization effort. As part of our future work, we would like to incorporate active learning techniques in order to filter out only those batches that need human inspection, while rest of the batches will be corrected automatically.